\documentclass{article}

\usepackage{PRIMEarxiv}

\usepackage[utf8]{inputenc} 
\usepackage[T1]{fontenc}    
\usepackage{url}            
\usepackage{booktabs}       
\usepackage{amsfonts}       
\usepackage{nicefrac}       
\usepackage{makecell}
\usepackage{microtype}      
\usepackage{lipsum}
\usepackage{fancyhdr}       
\usepackage{graphicx}       
\usepackage{amsmath}
\usepackage{tabularx}
\usepackage{bbm}
\usepackage{cite}
\usepackage{array}
\usepackage{authblk}
\graphicspath{{media/}}     
\usepackage{xcolor}

\usepackage[
  colorlinks=true,
  linkcolor=black,     
  citecolor=blue,     
  urlcolor=magenta     
]{hyperref}

\title{RAG: A Random-Forest-Based Generative Design Framework for Uncertainty-Aware Design of Metamaterials with Complex Functional Response Requirements}

\author[1]{Bolin Chen}
\author[1]{Dex Doksoo Lee}
\author[2]{Wei ``Wayne'' Chen}
\author[1*]{Wei Chen} 

\affil[1]{Department of Mechanical Engineering, Northwestern University, Evanston, IL 60208}
\affil[2]{J. Mike Walker ’66 Department of Mechanical Engineering, Texas A\&M University, College Station, TX 77843}
\affil[*]{Corresponding author: weichen@northwestern.edu} 

\begin{document}
\maketitle

\begin{abstract} 

Metamaterials design for advanced functionality often entails the inverse design on nonlinear and condition-dependent responses (e.g., stress-strain relation and dispersion relation), which are described by continuous functions. Most existing design methods focus on vector-valued responses (e.g., Young’s modulus and bandgap width), while the inverse design of functional responses remains challenging due to their high-dimensionality, the complexity of accommodating design requirements in inverse-design frameworks, and non-existence or non-uniqueness of feasible solutions. Although generative design approaches have shown promise, they are often data-hungry, handle design requirements heuristically, and may generate infeasible designs without uncertainty quantification. To address these challenges, we introduce a RAndom-forest-based Generative approach (RAG). By leveraging the small-data compatibility of random forests and reformulating the forward mapping in a discretization-invariant way, RAG enables data-efficient predictions of high-dimensional functional responses. During the inverse design, the framework estimates the likelihood of solutions conditioned on the design requirement that can be flexibly specified. The likelihood estimated through the ensemble quantifies the trustworthiness of generated designs while reflecting the relative difficulty across different requirements. The one-to-many mapping is addressed through single-shot design generation by sampling from the conditional likelihood. We demonstrate RAG on: 1) acoustic metamaterials with prescribed partial passbands/stopbands, and 2) mechanical metamaterials with targeted snap-through responses, using 500 and 1057 samples, respectively. Its data-efficiency is benchmarked against neural networks on a public mechanical metamaterial dataset with nonlinear stress-strain relations. Our framework provides a lightweight, trustworthy pathway to inverse design involving functional responses, expensive simulations, and complex design requirements, beyond metamaterials.

\end{abstract}

\section{Introduction} \label{introduction}

Metamaterials are artificially engineered materials whose extraordinary properties arise from their geometry rather than material composition \cite{yu2018mechanical, wang2020thermal, cummer2016controlling}. Through meticulous geometric design, they enable a wide range of applications---spanning thermal \cite{sklan2018thermal}, mechanical \cite{jiao2023mechanical}, acoustic \cite{liao2021acoustic}, and optical \cite{soukoulis2011past} regimes---that are unattainable with conventional materials. Designing metamaterials for target functionalities hinges on tailoring unit cell geometries to achieve desired responses. For instance, thermal cloaks can be realized by tessellating unit cells with tailored thermal conductivity tensors \cite{wang2023deep, da2024two}. Mechanical metamaterials exhibiting target deformation behaviors can be achieved through the spatial distribution of microstructures with local target stiffness \cite{wang2020deep}. Likewise, wave propagation in acoustic metamaterials can be manipulated by arranging unit cells with desired bandgap properties \cite{oudich2010propagation}.

\begin{figure}[!h]
\centering
\includegraphics[width=1.0\textwidth]{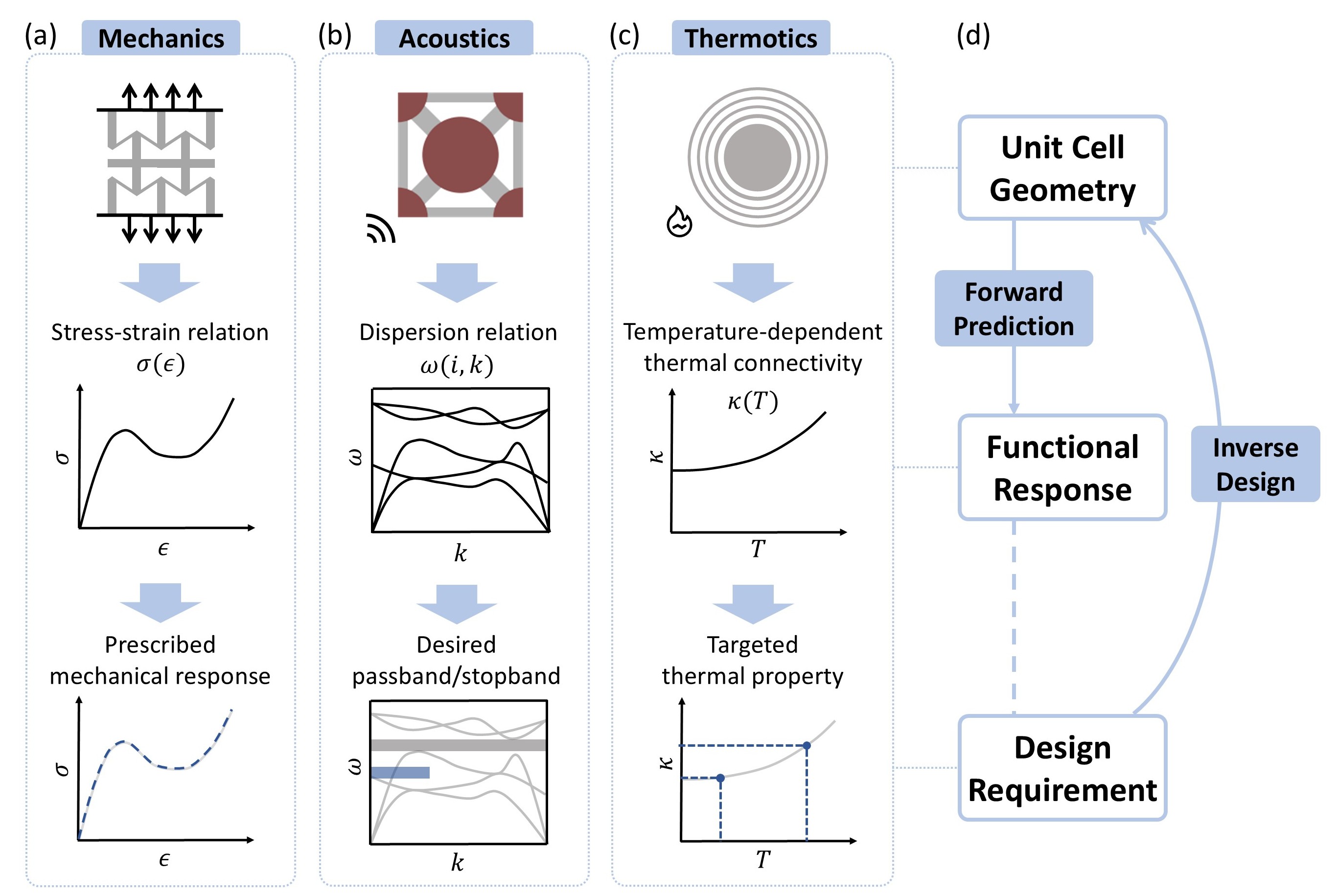}
\caption{Illustration of inverse design on functional responses across different classes of metamaterials. In each case, the middle panel shows the functional response, while the bottom panel illustrates the corresponding design requirement. (a) Mechanical metamaterials: The functional response is the stress--strain relation, where stress $\sigma$ is a function of strain $\epsilon$. The design requirement is a prescribed nonlinear mechanical behavior, represented by the blue dashed line. (b) Acoustic metamaterials: The functional response is the dispersion relation, where frequency $\omega$ is a function of band index $i$ and wave vector $k$. The design requirements are indicated by blue and gray shaded areas, representing the desired passbands and stopbands, respectively. (c) Thermal metamaterials: The functional response is the temperature-dependent thermal properties, where thermal connectivity $\kappa$ is a function of temperature $T$. The design requirement---high conductivity at high temperatures and low conductivity at low temperatures---is illustrated by the blue dashed line. (d) Schematic illustrating the relationship among unit-cell geometry, functional responses, and design requirements in metamaterials design. The form of the design requirements could vary across applications and is often complex to specify.}
\label{fig:functional response}
\end{figure}

The richness of unit-cell responses dictates the achievable system-level functionalities. Existing metamaterials design efforts primarily focus on vector-valued responses, such as stiffness tensors \cite{wang2020deep, liu2025machine, zheng2021controllable, ang2025deep, gao2023data, luan2023data, wang2025uncertainty}, bandgap ranges \cite{sun2025low, tran2025deep, liu2024data, wang2022design}, thermal conductivity tensors \cite{wang2023deep}, or their combination \cite{park2025investigating}, all of which can be represented as finite-dimensional vectors. Although vector-valued responses serve as an effective proxy for the underlying behaviors, they are often insufficient to fully characterize them. For example, stiffness characterizes only the linear elastic regime of the stress–strain relation but cannot describe nonlinear mechanical behaviors such as snap-through. In comparison, functional responses---expressed as infinite-dimensional functions\cite{mu2025guide, chen2025generative}---can describe nonlinear behavior with complex physical mechanisms (e.g., nonlinear stress--strain relation) or properties that vary under contextual factors (e.g., temperature-dependent thermal conductivity). Functional responses carry the low-level underlying behaviors from which vector-valued responses are derived. Thus, the design formulated with functional responses enables to unlock more advanced, sophisticated functionalities possibly unattainable by the other. For example, a soft actuator can be spatially and temporally programmed by distributing unit cells with prescribed stress--strain relations \cite{chai2024tailoring} (Fig.~\ref{fig:functional response} (a)). Directional noise filtering can be achieved by designing unit cells with desired dispersion relations that meet specific passband/stopband requirements \cite{valappil2025directional} (Fig.~\ref{fig:functional response} (b)). By tailoring the material's thermal conductivity across different temperatures, temperature-switchable cloaking and concentrating can be realized \cite{zhuang2022nonlinear} (Fig.~\ref{fig:functional response} (c)). ther demonstrations in the literature include advanced functionalities and applications such as shoe midsoles with tunable dynamic performance \cite{ha2023rapid}, lacrosse chest protectors and vibration-damping panels \cite{maurizi2025designing}, and energy absorption and dissipation \cite{zhao2025extreme, zeng2023inverse, giri2021controlled}.

Despite their promise, designing metamaterials with desired functional responses remains challenging due to the high-dimensionality of the responses and the non-uniqueness or non-existence of design solutions. To handle the infinite-dimensional nature of functions, a common practice is to discretize them into high-dimensional vector \cite{chai2024tailoring, zhang2024deep}. Such high dimensionality, often coupled with the nonlinearity of underlying structure–property relations, not only makes the evaluation of functional responses computationally expensive but also complicates the mapping between the design and response spaces. Moreover, a given design requirement on the functional response may admit many feasible designs if it is relatively easy to satisfy, or none if it is inherently stringent \cite{bastek2022inverting, mu2025guide}. In particular, design requirements imposed on functional responses are expressed in various forms: they may specify exact response values (e.g., an entire stress--strain curve shown in Fig.~\ref{fig:functional response} (a)) or only enforce certain characteristics (e.g., acoustic bandgap properties of the dispersion relation shown in Fig.~\ref{fig:functional response} (b)). Such complexity in requirement specification further challenges the inverse design.

Conventional optimization methods, such as topology optimization \cite{sigmund2013topology}, have been widely used for the metamaterials design involving functional responses \cite{zhao2025extreme, xu2025inverse, cool2025metamaterial, wu2023topology, azizi2025lattice}. In these approaches, the design requirements are formulated as objective functions, e.g., minimizing the discrepancy between the targeted stress--strain relation and the obtained one \cite{zhao2025extreme, xu2025inverse}, and design solutions are obtained through iterative evaluations of structure--property relations. Gradient-based topology optimization has proven effective for inverse design of acoustic bandgaps \cite{cool2025metamaterial, wu2023topology}, topological states\cite{azizi2025lattice}, and nonlinear stress–-strain relations \cite{zhao2025extreme, xu2025inverse}. However, analytical sensitivity analysis applies only to analytically differentiable objectives, limiting flexibility in design requirement specification. To bypass sensitivity derivation, non-gradient-based optimization methods such as genetic algorithms have been employed \cite{zeng2023inverse, zhang2021realization, zhang2023ultra}, providing greater flexibility in handling complex design requirements that are difficult to formulate as differentiable objectives. Nonetheless, they tend to require a large number of evaluations of functional responses, which are often time-consuming. Iterative optimications in general are sensitive to the initial guess, which limits their ability to efficiently explore diverse design solutions. 

Data-driven design \cite{lee2024data, zheng2023deep} offers a promising alternative for inverse design of metamaterials with functional responses. Central to the framework is a surrogate forward model that can be trained for rapid, on-the-fly prediction without costly evaluations \cite{lee2024data}, then be integrated with optimization at the downstream to accelerate the search process \cite{deng2022inverse, tang2025data, baali2023design, cao2025inverse}, or combined with another machine learning model supporting one-shot, iteration-free inverse design \cite{ha2023rapid, chai2024tailoring, zhang2024deep, liang2025demand, jiang2022dispersion}. More recently, conditional deep generative models including generative adversarial networks \cite{gurbuz2021generative, yan2025inverse}, diffusion model \cite{bastek2023inverse, vlassis2023denoising, liu2025toward}, variational autoencoders \cite{wang2021functional}, have been applied to generate design candidates conditioned on functional responses, such as stress--strain relations and acoustic dispersion relations. Since generative models are inherently stochastic, they can more effectively accommodate one-to-many mappings than deterministic models \cite{ma2019probabilistic}. The diversity of design solutions offers the flexibility to consider additional design targets beyond the primary requirement.

Nonetheless, handling the high dimensionality of functional responses with machine learning tends to require large datasets---often on the order of $10^3$-$10^5$ samples \cite{chai2024tailoring, ha2023rapid, zhang2024deep}. When the computational cost of evaluating functional responses is high, acquiring such large datasets becomes impractical. Furthermore, most generative models are conditioned directly on the response rather than on the requirement \cite{gurbuz2021generative, yan2025inverse, bastek2023inverse, vlassis2023denoising, wang2021functional, ma2019probabilistic}, only supporting requirements that are tied to specific responses (e.g., the entire dispersion relations must be provided as the conditioning response for acoustic bandgap inverse design \cite{zhang2024deep}). In addition, because existing generative design approaches provide no uncertainty estimate in the conditional inference, we cannot assess the trustworthiness of the generated design solutions. This is particularly problematic given the potential non-existence or non-uniqueness of feasible solutions: without uncertainty assessment across them, the method may produce designs that are actually infeasible, especially when the requirement lies outside the training distribution or cannot be satisfied within the predefined design space \cite{mu2025guide}.

To date, most data-driven design approaches have employed neural networks (NN) \cite{lee2024data, zheng2023deep}. While their strong expressive power facilitates the accurate prediction of functional responses, NNs are data-hungry, require extensive hyperparameter tuning, and often demand resource-intensive training. In contrast, random forests---another widely adopted machine learning method---are relatively easy to construct, exhibiting greater stability to training parameter variations \cite{rodriguez2015machine}. They are particularly advantageous in small-data regimes, which can reduce the data demand in metamaterials design, where simulations to obtain functional responses are computationally intensive. Furthermore, their ensemble structure naturally supports uncertainty quantification of the model predictions \cite{zhang2022uncertainty}. Random forests have proven effective for predicting functional response such as stress--strain relations \cite{he2025customizable} and optical transmission spectra \cite{chouhan2024machine}, but their inherent capability for uncertainty quantification has rarely been utilized for inverse design. Our previous work \cite{chen2025generative} leverages the ensemble structure of random forests to provide confidence estimates for generated solutions. The data efficiency of this approach was demonstrated through spectral inverse design of acoustic metamaterials and optical metasurfaces, requiring fewer than 250 samples in each example. Nevertheless, this framework was limited to simplified, qualitative functional responses---such as binary representation of spectral behavior in acoustics and optics---and did not address fully quantitative ones. 

In this work, we propose a RAndom-forest-based Generative design framework (RAG), which can tackle the inverse design of quantitative, high-dimensional functional responses with complex requirement specification, small-data compatibility, and uncertainty quantification over generated designs. A random forest is trained for forward mapping, where each decision tree predicts the entire functional response from a given unit-cell geometry. Leveraging this full functional-response information, complex design requirements can be specified on either specific response values or characteristics extracted from the response. Each tree can then classify whether its predicted response satisfies that requirement. Aggregating the votes across trees in the random forest yields a likelihood estimate conditioned on the requirement, indicating the model's confidence that a given design can achieve the on-demand functional response. The one-to-many mapping can be resolved through iteration-free, single-shot design generation by sampling the design space based on the conditional likelihood distribution. 


In the context of metamaterial inverse functional response design, the contributions of RAG are as follows:
\begin{enumerate}
\item \textbf{Data-efficiency.} RAG can predict high-dimensional functional responses accurately using small datasets, which is beneficial given the high computational cost of evaluating functional responses. 
\item \textbf{Handling complex requirement specification.} RAG can accommodate complex forms of design requirements on functional response, providing greater freedom in inverse functional response design.
\item \textbf{Uncertainty quantification over generated solutions.} RAG can provide predictive uncertainty when generating design solutions. The uncertainty information can help prevent accepting designs with low confidence in meeting requirements, improving trustworthiness especially when requirements lie outside of distribution.
\end{enumerate}

To validate the proposed framework, we investigate two functional response inverse design tasks: (1) acoustic metamaterials with prescribed partial passband/stopband behavior, and (2) mechanical metamaterials with targeted snap-through responses \cite{chai2024tailoring}. Both studies are conducted with substantially smaller datasets than existing works\cite{ha2023rapid, chai2024tailoring, deng2022inverse, zhang2024deep}—500 samples for the acoustic case and 1057 samples for the mechanical case. The design requirements in both tasks are difficult to formulate in differentiable objectives thus challenging to solve using topology optimization. The results demonstrate that RAG can efficiently handle high-dimensional functional response and complex design requirements with small data. The conditional likelihood, which indicates the model’s uncertainty, not only reflects the trustworthiness of generated designs for a given requirement, but also reveals the relative difficulty across different requirements, thereby helping to prevent the generation of infeasible designs. The data efficiency is validated on a public dataset by comparing with a NN-based design framework \cite{chai2024tailoring}. Overall, this lightweight and uncertainty-aware design framework provides a fast and easy-to-implement approach for inverse functional response design of metamaterials in the small-data regime.

The remainder of the paper is organized as follows. Sec.~\ref{Method} introduces the mathematical formulation of the functional response inverse design problem and the architecture of RAG. In Sec.~\ref{Results}, we validate RAG's small-data compatibility and uncertainty quantification capability on acoustic metamaterials design (Sec.~\ref{Design Acoustic Metamaterials with On-Demand Partial Passbands/Stopbands}) and mechanical metamaterials design (Sec.~\ref{Design Mechanical Metamaterials with Target Snap-Through Response}). Discussions and conclusions are stated in Sec.~\ref{Discussion} and Sec.~\ref{Conclusion}, respectively.

\section{Methods} \label{Method}

In this section, we first provide a formal description of the inverse functional response design problem of our interest and discuss the associated key challenges, including the high-dimensionality of the functional responses, complex design requirements, and potential non-existence or non-uniqueness of design solutions (Sec.~\ref{Problem Formulation of Inverse Functional Response Design}). To address these challenges, we introduce the RAG framework (Sec.~\ref{RAG framework}), which enables data-efficient forward prediction of high-dimensional functional responses (Sec.~\ref{Forward Prediction with Uncertainty Quantification}) and uncertainty-informed inverse design under complex requirement specifications (Sec.~\ref{Uncertainty-Informed Inverse Design}).

\subsection{Problem Formulation of Inverse Functional Response Design}
\label{Problem Formulation of Inverse Functional Response Design}

\begin{figure}[!h]
\centering
\includegraphics[width=0.98\textwidth]{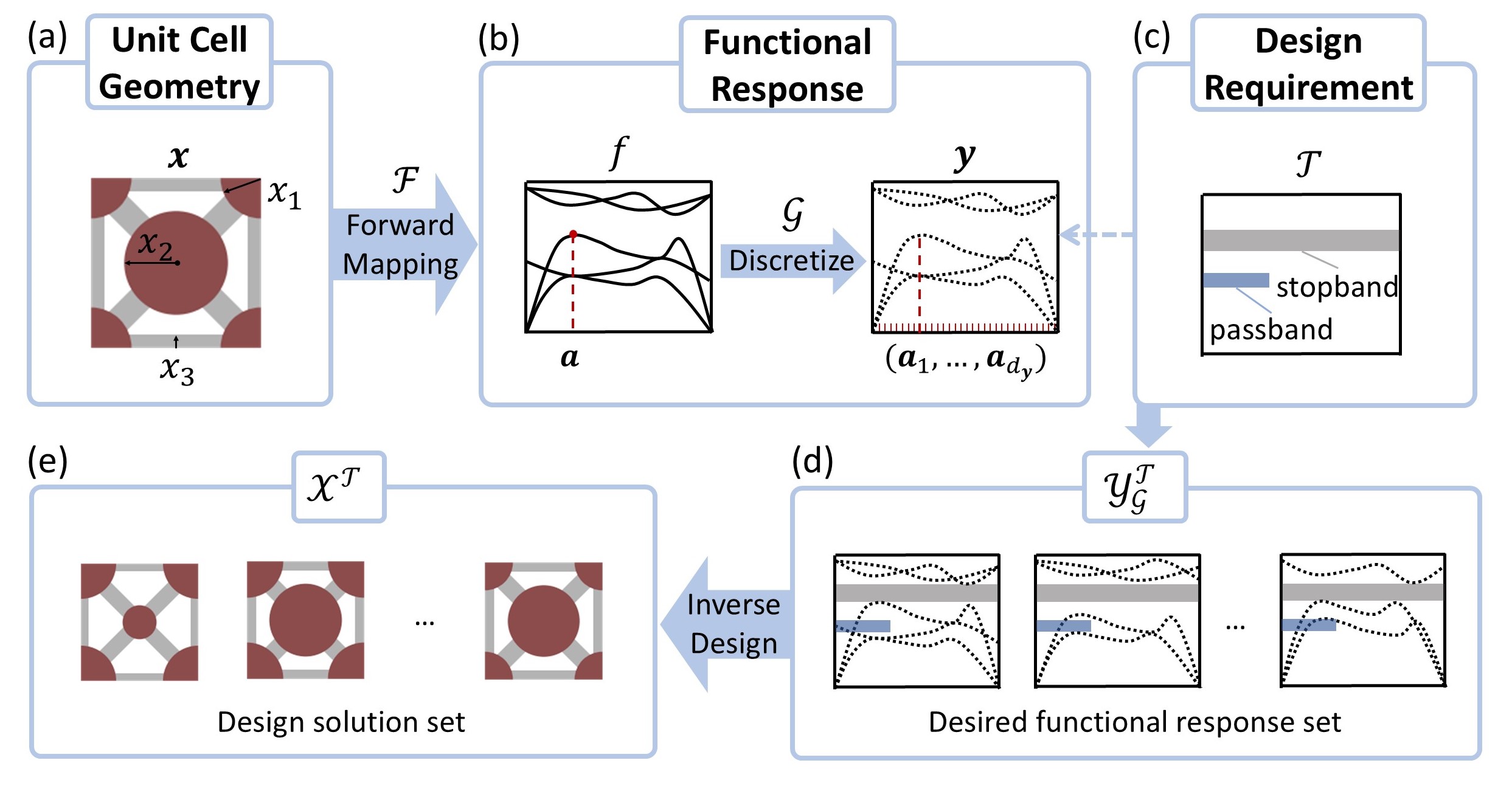}
\caption{Formulation of forward mapping and inverse design, illustrating the relationship between (a) the unit cell geometry $\mathbf{x}$, (b) corresponding functional response $f$ and its discrete form $\mathbf{y}$, (c) design requirement $\mathcal{T}$, (d) desired functional response set $\mathcal{Y}^\mathcal{T}_\mathcal{G}$, and (e) design solution set $\mathcal{X}^\mathcal{T}$.}
\label{fig:problem formulation}
\end{figure}

The forward mapping from a metamaterial unit cell design to its functional response shown in Fig.~\ref{fig:problem formulation} (a) and (b) can be expressed as 
\begin{equation} 
\mathcal{F}(\mathbf{x}) = f,
\label{eq:mapping}
\end{equation}
where $\mathcal{F}: \Omega_{\mathbf{x}} \to 
\{ f : \Omega_{\mathbf{a}} \to \mathbb{R} \}$. $\mathbf{x}$ is the vector of design variables with dimension $d_{\mathbf{x}}$ representing the unit cell geometry. The corresponding design space is $\Omega_{\mathbf{x}} \subset \mathbb{R}^{d_{\mathbf{x}}}$. $f$ denotes the functional response of interest, defined over the domain $\Omega_{\mathbf{a}} \subseteq \mathbb{R}^{d_{\mathbf{a}}}$ of the query point $\mathbf{a}$. For example, $f$ can represent a stress--strain relation $\sigma(\epsilon)$ (where query point $\mathbf{a}$ is the strain $\epsilon$) or dispersion relation $\omega(i,k)$ (where $\mathbf{a} = (i, k)$, with $i$ and $k$ denoting the band index and wave vector, respectively). $\Omega_{\mathbf{a}}$ specifies the corresponding range of interest.

Since the functional response $f$ in Eq.~\ref{eq:mapping} is infinite-dimensional and difficult to directly handle, we discretize it over a sequence of query points $(\mathbf{a}_1, \ldots, \mathbf{a}_{d_{\mathbf{y}}})$ in its domain $\Omega_{\mathbf{a}}$, as shown in Fig.~\ref{fig:problem formulation} (b). Evaluating $f$ at these query points defines a discretization operator
\begin{equation}
 \mathcal{G}(f) = \mathbf{y},
\label{eq:discretization operator}
\end{equation}
where $\mathcal{G}:   
\{ f : \Omega_{\mathbf{a}} \to \mathbb{R} \} \to \mathbb{R}^{d_{\mathbf{y}}}$. $\mathbf{y}$ is the discretized functional response, with components $y_q = f_{\mathbf{x}}(\mathbf{a}_q), q=1,\ldots,d_{\mathbf{y}}$. The specific definition of the discretization operator $\mathcal{G}$ depends on the sampling strategy over $\Omega_{\mathbf{a}}$, which must ensure $\mathbf{y}$ preserves the necessary information of the original infinite-dimensional functional response $f$. In practice, highly nonlinear areas of $f$ necessitate a higher density of query points for an accurate representation. Therefore, more complex functional responses often require more query points for discretization, leading to high dimensionality. Without loss of generality, we consider uniform sampling for simplicity throughout this work. Specifically, the domain $\Omega_\mathbf{a}$ is sampled uniformly, where each dimension $a_j$ is discretized into $n_j$ equidistant points $(a_{j,1}, \ldots, a_{j,n_j}), j=1,\ldots, d_{\mathbf{a}}$. Consequently, the total dimension of $\mathbf{y}$ is  $d_{\mathbf{y}} = \prod_{j=1}^{d_{\mathbf{a}}} n_j$. This indicates that the dimension of $\mathbf{y}$ scales exponentially with the dimensionality of $\mathbf{a}$, since each additional dimension multiplies the number of required query points. Combining Eqs.~\ref{eq:mapping} and ~\ref{eq:discretization operator}, the discretized forward mapping can be obtained as
\begin{equation}
(\mathcal{G}\circ \mathcal{F})(\mathbf{x}) = \mathbf{y},
\label{eq:discretized mapping}
\end{equation}
where $\mathcal{G}\circ \mathcal{F}:\;
\Omega_{\mathbf{x}}\to\mathbb{R}^{d_{\mathbf{y}}}$. 

The design requirement defines which functional responses are considered desirable. To facilitate determining whether a response is desired, the requirements are imposed on the discrete form of functional response $\mathbf{y}$. We define the desired response set $\mathcal{Y}^{\mathcal{T}}_{\mathcal{G}}$ as the set of all discretized responses $\mathbf{y}$, constructed under $\mathcal{G}$, that satisfy the requirement $\mathcal{T}$ (as shown in Fig.~\ref{fig:problem formulation} (d)). A straightforward type of $\mathcal{T}$ is to prescribe an target functional response $\mathbf{y}^*$, such as a nonlinear stress--strain relation \cite{wang2024diffmat, liu2025toward, ha2023rapid, zhang2025generative, mu2025guide}, in which case $\mathcal{Y}^{\mathcal{T}}_{\mathcal{G}} = \{\mathbf{y}^*\}$. If some deviation from the target is acceptable, the requirement then admits a set of responses within a tolerance band $\mathcal{Y}^{\mathcal{T}}_\mathcal{G} = \{\mathbf{y}|\ |y_i - y_i^*| \le \delta_i,\; \forall i = 1,\dots,d_{\mathbf{y}}\}$, where $\delta_i$ denotes the allowable deviation in each component. In other cases, the requirement concerns certain characteristics of the functional response rather than its full profile. Typical examples include specifying plateau stress or absorbed energy in mechanical metamaterials \cite{ang2025deep, chai2024tailoring}, or targeting a specific full bandgap with the lower bound $\omega_l$ and upper bound $\omega_u$ in acoustic metamaterials \cite{bao2025phononic} (as illustrated by the gray shaded regions in Fig.~\ref{fig:problem formulation} (c) and (d)), which can be expressed as $\mathcal{Y}^{\mathcal{T}}_\mathcal{G} = \{\mathbf{y}\mid y_i \not \in [\omega_l, \omega_u],\; \forall i = 1,\dots,d_{\mathbf{y}}\}$. Under such characteristic-related requirements, the full functional response is not uniquely determined, meaning that responses with distinct overall patterns may still be feasible. Consequently, the elements in $\mathcal{Y}^\mathcal{T}_\mathcal{G}$ can exhibit substantial diversity, as illustrated in Fig.~\ref{fig:problem formulation} (d). This, in turn, often leads to a diverse set of feasible design solutions.

When the requirement $\mathcal{T}$ is specified, inverse design aims to find the corresponding design solution set $\mathcal{X}^{\mathcal{T}}$, which consists of all designs in the design space $\Omega_{\mathbf{x}}$ whose functional responses satisfy $\mathcal{T}$, as shown in Fig.~\ref{fig:problem formulation} (e). Based on the discretized forward mapping in Eq.~\ref{eq:discretized mapping}, $\mathcal{X}^{\mathcal{T}}$ is defined as 
\begin{equation}
\mathcal{X}^{\mathcal{T}}
=
\left\{
\mathbf{x} \in \Omega_{\mathbf{x}}\mid
(\mathcal{G} \circ \mathcal{F})(\mathbf{x}) \in \mathcal{Y}^{\mathcal{T}}_\mathcal{G}
\right\}.
\label{eq:solution set}
\end{equation}
It is worth noting that the solution set $\mathcal{X}^{\mathcal{T}}$ in Eq.~\ref{eq:solution set} may contain multiple feasible designs if $\mathcal{T}$ is easy to satisfy, or may be empty if $\mathcal{T}$ is tough to reach within the given design space. This non-uniqueness or potential non-existence of design solutions introduces significant challenges in inverse functional design.

In summary, the high-dimensional nature of functional responses, the complexity of design requirements, and the non-uniqueness or potential non-existence of design solutions render the inverse design process highly challenging. In Sec.~\ref{RAG framework}, we will present the proposed RAG framework and demonstrate how it systematically addresses these challenges.

\subsection{The RAG Framework}
\label{RAG framework}

\subsubsection{Forward Prediction with Uncertainty Quantification}
\label{Forward Prediction with Uncertainty Quantification}


According to Eq.~\ref{eq:solution set}, identifying the solution set $\mathcal{X}^{\mathcal{T}}$ requires building a surrogate model for the forward mapping $\mathbf{y} = (\mathcal{G} \circ \mathcal{F})(\mathbf{x})$. Here $\mathcal{F}$ represents the physical mapping which can not be changed, while the discretization operator $\mathcal{G}$ is manually specified based on the nonlinearity of the response. The most straightforward way to construct the surrogate model is to take $\mathbf{x}$ as input and $\mathbf{y}$ as output. However, learning a mapping to high-dimensional output spaces suffers from the curse of dimensionality, necessitating prohibitive amounts of data for effective generalization. To mitigate this issue, existing approaches often rely on (i) dimensionality reduction methods (e.g., principal component analysis \cite{deng2022inverse, nakarmi2024predicting}) to obtain a low-dimensional form of $\mathbf{y}$, (ii) manual parameterization $\mathbf{y}$ into a compact form \cite{ha2023rapid}, or (iii) partitioning $\mathbf{y}$ into several segments and train separate models for each \cite{zhang2024deep}. Despite their effectiveness, these methods still require large datasets and complex model architectures \cite{chai2024tailoring, ha2023rapid, zhang2024deep}. Furthermore, because the learned mapping $(\mathcal{G} \circ \mathcal{F})$ is tied to a fixed discretization $\mathcal{G}$, disabling to make predictions at arbitrary query points inside $\Omega_{\mathbf{a}}$. 

\begin{figure}[!h]
\centering
\includegraphics[width=0.98\textwidth]{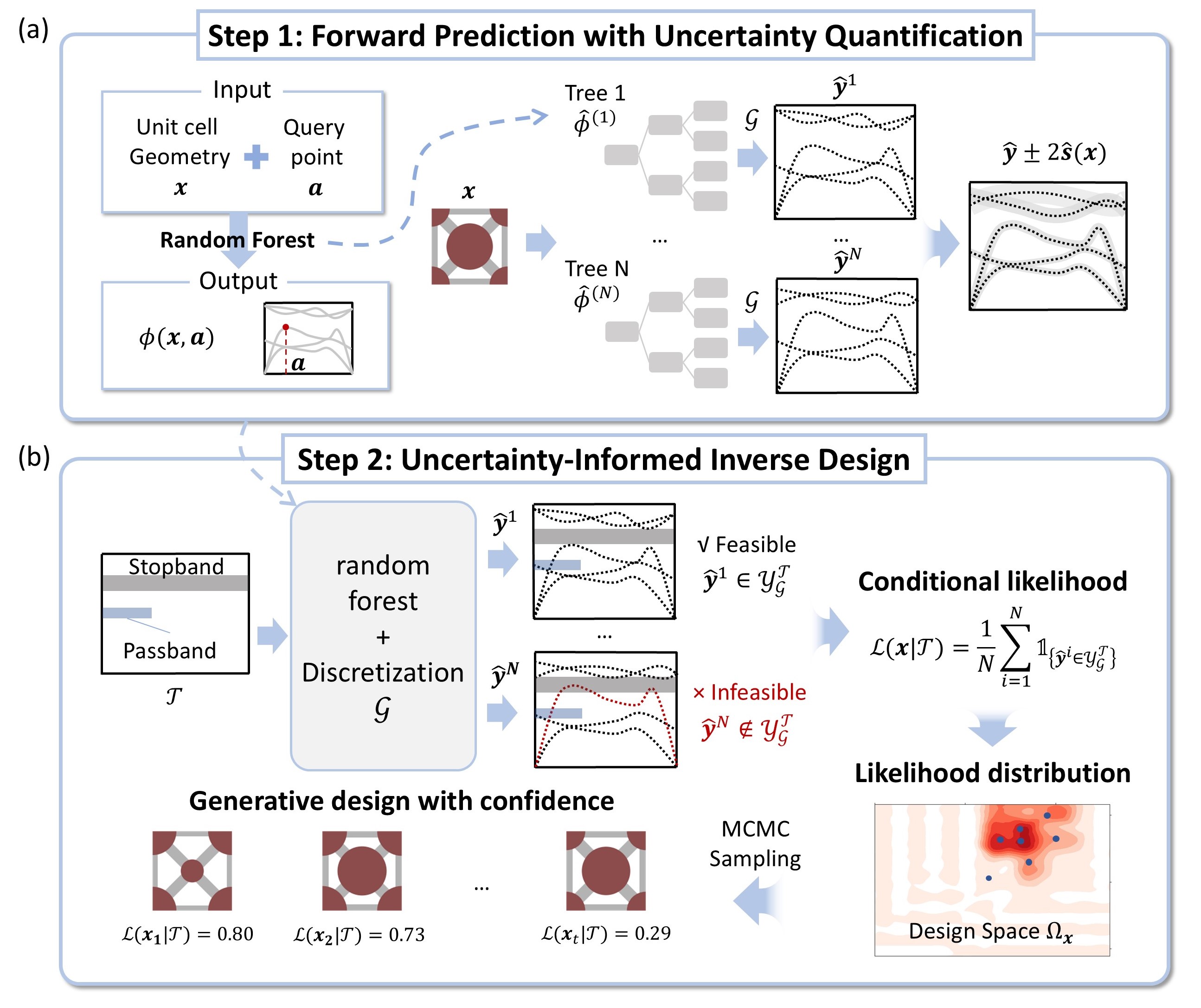}
\caption{Illustration of the RAG framework. (a) Step 1: forward prediction of $\mathbf{y}$ from $\mathbf{x}$ with uncertainty quantification (c) Step 2: uncertainty-informed inverse design given a design requirement $\mathcal{T}$.}
\label{fig:RAG_framework}
\end{figure}

These two issues can be addressed by reformulating the original mapping in Eq.~\ref{eq:mapping} as, 
\begin{equation} 
\phi(\mathbf{x}, \mathbf{a})
= \mathcal{F}(\mathbf{x})(\mathbf{a}),
\label{eq:uncurried}
\end{equation}
where $\phi : \Omega_{\mathbf{x}} \times \Omega_{\mathbf{a}} \to \mathbb{R}$. As shown in Fig.~\ref{fig:RAG_framework} (a), by taking both the design $\mathbf{x}$ and the query point $\mathbf{a}$ as input, the learned surrogate $\hat{\phi}$ can predict the response at arbitrary location $\mathbf{a} \in \Omega_{\mathbf{a}}$. This discretization invariance allows the functional response to be transferred across different discretizations. For a given discretization operator $\mathcal{G}$, the corresponding discrete response is obtained as $\mathbf{y}= (\mathcal{G}\circ \phi)(\mathbf{x})$, with each component given by $y_q= \phi(\mathbf{x}, \mathbf{a}_q)$. Additionally, since the output dimensionality is one, the surrogate can be trained with less data and learn the forward mapping more efficiently \cite{yang2025deep}.

In RAG, random forest is implemented to approximate the mapping in Eq.~\ref{eq:uncurried}. Random forests are an ensemble learning method that constructs multiple decision trees trained on the same task and aggregates their predictions to improve accuracy. Compared with NNs, random forests can work effectively on small datasets, require less hyperparameter tuning, and train more efficiently. Moreover, they naturally handle mixed data types, as $\mathbf{x}$ may contain both numerical and categorical variables. After training the random forest with $N$ trees, each tree $\hat{\phi}^{(n)}$ predict its own discrete functional response $\hat{\mathbf{y}}^{(n)} = (\mathcal{G}\circ \hat{\phi}^{(n)})(\mathbf{x}), n=1,\ldots,N$. The ensemble $\hat{\phi}$ is then defined as the average of all trees $\hat{\phi}(\mathbf{x},\mathbf{a}) =  \frac{1}{N}\sum_{n=1}^{N}\hat{\phi}^{(n)}(\mathbf{x},\mathbf{a})
$, and its corresponding discrete prediction is
$\hat{\mathbf{y}}=(\mathcal{G}\circ \hat{\phi})(\mathbf{x})$.

Beyond improving accuracy through averaging, the ensemble structure of random forests also provides predictive uncertainty, which arises from the variability among individual tree outputs \cite{ling2017high, zhang2022uncertainty}, as shown in Fig.~\ref{fig:RAG_framework} (a). The variance of the tree predictions reflects the degree of disagreement among trees: small variance indicates consistent, confident predictions, while large variance suggests greater uncertainty. For computational simplicity, we ignore the correlation among tree outputs and estimate the predictive uncertainty using the sample variance: $\hat{\mathbf{s}}^2(\mathbf{x}) =
\frac{1}{N}\sum_{n=1}^{N}
\big((\mathcal{G}\circ\hat{\phi}^{(n)})(\mathbf{x})
- (\mathcal{G}\circ\hat{\phi})(\mathbf{x})\big)^{\circ 2}
$, where $(\cdot)^{\circ 2}$ indicates element-wise squaring. The resulting $\hat{\mathbf{s}}^{2}(\mathbf{x}) \in \mathbb{R}^{d_{\mathbf{y}}}$ is a vector-valued uncertainty estimate, with each entry $\hat{s}^{2}_{q}(\mathbf{x})$ quantifying the predictive variance at the corresponding query point $\mathbf{a}_{q}$. This provides an estimate of the model’s confidence, enabling us to assess the trustworthiness of the predictions used in the inverse design process.


\subsubsection{Uncertainty-Informed Inverse Design}
\label{Uncertainty-Informed Inverse Design}

As illustrated in Fig.~\ref{fig:problem formulation} (c) and (d), when the design requirement $\mathcal{T}$ is specified, the desired functional response set $\mathcal{Y}^{\mathcal{T}}_{\mathcal{G}}$ can be defined. For a given design $\mathbf{x}$, each decision tree $\hat{\phi}^{(n)}$ in the random forest determines whether $\mathbf{x}$ satisfies the requirement (shown in Fig.~\ref{fig:RAG_framework} (b)). Based on the proportion of trees that classify $\mathbf{x}$ as feasible, we can estimate a conditional likelihood that quantifies how likely $\mathbf{x}$ satisfies $\mathcal{T}$:
\begin{equation}
\mathcal{L}(\mathbf{x}\mid\mathcal{T}) = \frac{1}{N}\sum_{n=1}^{N}\mathbbm{1}_{\{(\mathcal{G}\circ\hat{\phi}^{(n)})(\mathbf{x})\in \mathcal{Y}^{\mathcal{T}}_{\mathcal{G}}\}},
\label{eq:likelihood}
\end{equation}
where $\mathbbm{1}$ is the indicator function that returns 1 when the predicted response $\hat{\mathbf{y}}^{(n)} = (\mathcal{G}\circ\hat{\phi}^{(n)})(\mathbf{x})$ from tree $n$ belongs to the desired response set $\mathcal{Y}^{\mathcal{T}}_{\mathcal{G}}$ and 0 otherwise.
The conditional likelihood in Eq.~\ref{eq:likelihood} reflects the model’s predictive confidence. The more trees voting that $\mathbf{x}$ is feasible, the higher the model’s confidence that $\mathbf{x}$ satisfies the requirement. 

We remark that computing $\mathcal{L}(\mathbf{x}\mid\mathcal{T})$ in Eq.~\ref{eq:likelihood} relies solely on forward prediction, therefore no additional machine-learning module is required to learn an inverse mapping. Moreover, the likelihood is conditioned directly on $\mathcal{T}$. Existing generative models often learn the conditional distribution $p(\mathbf{x}|\mathbf{y})$, which condition directly on the responses $\mathbf{y}$ rather than requirement $\mathcal{T}$ \cite{gurbuz2021generative, yan2025inverse, bastek2023inverse, vlassis2023denoising, wang2021functional, ma2019probabilistic}. As a result, these methods generally require manually selecting a representative response \( \mathbf{y} \in \mathcal{Y}^{\mathcal{T}}_\mathcal{G} \) and providing it to the generative model to generate design solutions. This can be challenging when the desired functional response cannot be explicitly defined. Moreover, such heuristic selection also inevitably introduces bias that limits solution diversity, especially when $\mathcal{Y}_\mathcal{G}^\mathcal{T}$ contains highly diverse response patterns, as illustrated in Fig.~\ref{fig:problem formulation}. In comparison, in our framework, as long as the requirement $\mathcal{T}$ can be formulated as constraints defining $\mathcal{Y}^{\mathcal{T}}_{\mathcal{G}}$, the corresponding conditional likelihood can be estimated directly for design generation. This provides substantial flexibility in how design requirements can be specified, as further demonstrated in Secs.~\ref{Design Acoustic Metamaterials with On-Demand Partial Passbands/Stopbands} and ~\ref{Design Mechanical Metamaterials with Target Snap-Through Response}.

Based on the likelihood function, design solutions satisfying the requirement are obtained in a single shot by sampling from the likelihood distribution $\mathcal{L}(\mathbf{x}\mid\mathcal{T})$ over the design space. Because $\mathcal{L}(\mathbf{x}\mid\mathcal{T})$ may exhibit a complex shape depending on $\mathcal{T}$, we adopt the Metropolis-Hastings algorithm \cite{hastings1970monte}---a widely used Markov Chain Monte Carlo (MCMC) method---to generate samples from this likelihood. In each iteration, the algorithm proposes a new state according to a specified proposal distribution and decides whether to accept it based on the Metropolis-Hastings acceptance probability. This process constructs a Markov chain whose stationary distribution coincides with $\mathcal{L}(\mathbf{x}\mid\mathcal{T})$.

At iteration $k$, the new proposed state $\mathbf{x}_{k+1}$ is drawn from:
\begin{equation}
\mathbf{x}_{k+1} \sim \mathcal{N}\!\left(
\mathbf{x}_k,\,
\mathrm{diag}(\boldsymbol{\sigma}^2)
\right),
\label{eq:numerical_proposal}
\end{equation}
where the proposal standard deviation is defined as
\begin{equation}
\boldsymbol{\sigma}
= c_0 d_{\mathbf{x}}^{-1/2}
\lvert \mathbf{u} - \mathbf{l} \rvert,
\label{eq:sigma_def}
\end{equation}
with $\mathbf{l}$ and $\mathbf{u}$ denoting the lower and upper bounds of the design variables, respectively. Scaling the proposal standard deviation by the variable ranges allows the proposal step size to adapt naturally to the design space. $c_0$ serves as a scaling parameter controlling the overall step size. Since the Gaussian proposal is symmetric, the acceptance probability can be expressed as
\begin{equation}
\alpha(\mathbf{x}_{k},\mathbf{x}_{k+1}) = \min (1, \frac{
\mathcal{L}(\mathbf{x}_{k+1} \mid \mathcal{T})}{L(\mathbf{x}_k\mid\mathcal{T})}).
\label{eq:acceptance ratio]}
\end{equation}
Designs with higher likelihood values have a greater probability of being sampled, meaning that the RAG framework favors generating design solutions in regions where the surrogate model is more confident. In addition, the likelihood associated with each generated solution provides a quantitative indicator of its trustworthiness. If the requirement $\mathcal{T}$ is inherently difficult---or even impossible---to satisfy within the design space $\Omega_{\mathbf{x}}$, it will lead to a relatively low likelihood distribution across the entire domain. Thus this likelihood distribution can serve as a proxy that quantifies task difficulty across different design requirements, which is often not available in inverse design with deep generative modeling.

\section{Results}
\label{Results}

In this section, we evaluate the performance of RAG on the inverse design of dispersion relations in acoustic metamaterials (Sec.~\ref{Design Acoustic Metamaterials with On-Demand Partial Passbands/Stopbands}) and stress--strain relations in mechanical metamaterials (Sec.~\ref{Design Mechanical Metamaterials with Target Snap-Through Response}). For both tasks, the random forest is configured with 100 trees, setting the minimum number of samples for node splitting to 2 and for leaf nodes to 1. Gini impurity \cite{disha2022performance} is used as the node-splitting criterion. To account for different levels of nonlinearity, the maximum tree depth is adjusted for each task to balance predictive capability and computational efficiency.

\subsection{Design Acoustic Metamaterials with On-Demand Partial Passbands/Stopbands}
\label{Design Acoustic Metamaterials with On-Demand Partial Passbands/Stopbands}

\subsubsection{Problem Statement}
\label{Problem Statement}

Capitalizing on the unconventional capability to control propagation of acoustic/elastic waves in specified frequency ranges or directions, acoustic metamaterials have enabled a wide range of applications in vibration control \cite{fang2024advances}, signal processing \cite{zhang2021realization}, and energy harvesting \cite{akbari2024defect}. In general, designing acoustic metamaterials with targeted functionalities hinges on inversely designing unit cells that realize a desired dispersion relation \cite{kazemi2023drawing, ogren2024gaussian}. Therefore, the functional response $f$ in Eq.~\ref{eq:mapping} is a frequency function $\omega(i, \mathbf{k})$, with the query point $\mathbf{a} = (i, \mathbf{k})$, where $i$ denotes the band index and $\mathbf{k}$ denotes the wave vector. Although dispersion relations comprehensively describe the wave properties supported by a lattice of unit cell, their high dimensionality often leads existing works to reduce them to a few frequency-based features (e.g., full bandgap range) \cite{liu2023deep, chang2024demand, chen2025generative}. This simplification limits the potential of acoustic metamaterials for more sophisticated wave manipulation. In this work, we demonstrate that our RAG can achieve full dispersion relation inverse design through an acoustic metamaterial case study. 

\begin{figure}[!h]
\centering
\includegraphics[width=1\textwidth]{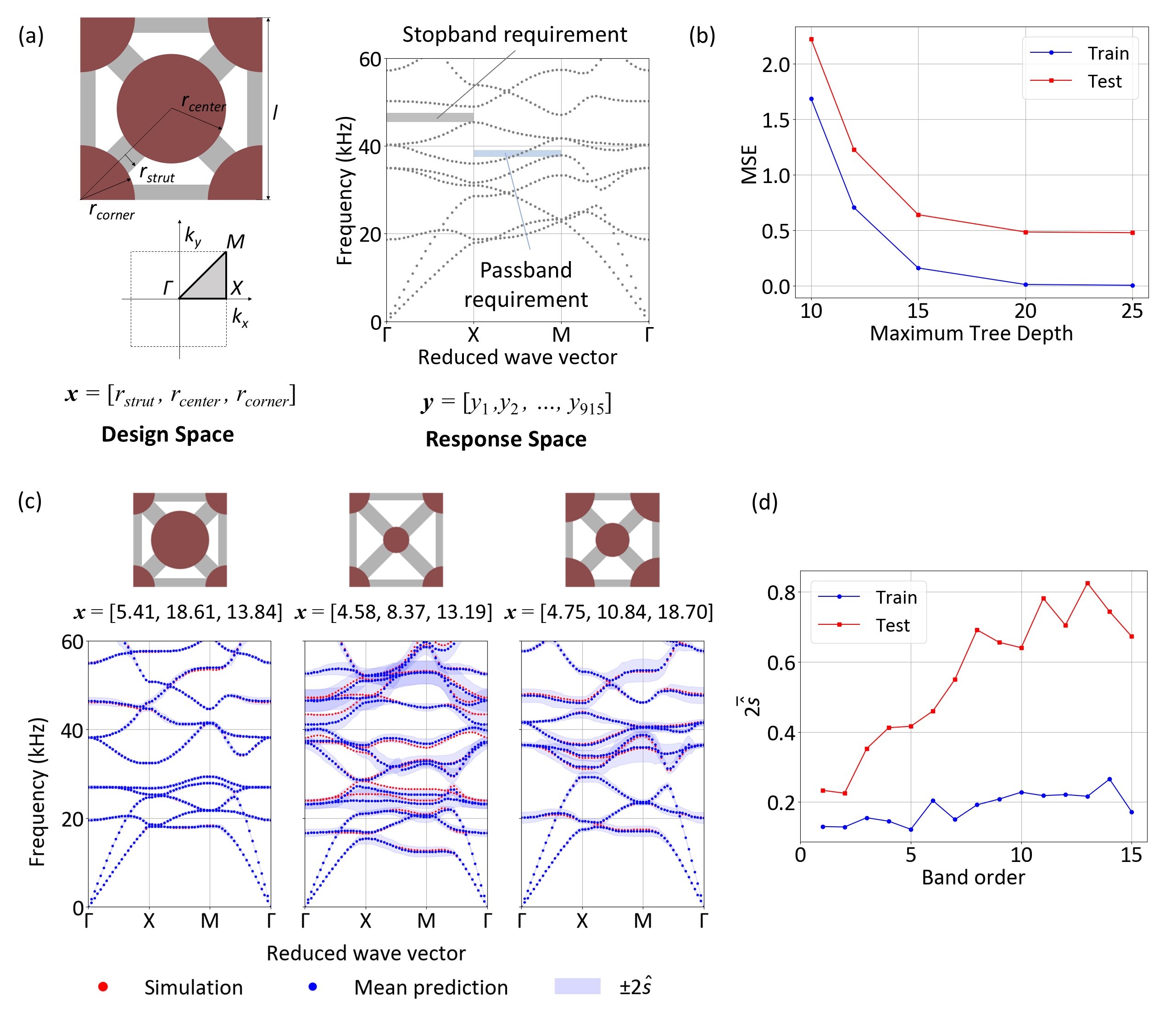}
\caption{(a) Design space and response space in the 2d acoustic metamaterials in the case study. The unit cell structure is specified by three geometric parameters $\mathbf{x} = [r_\text{strut}, r_\text{center}, r_\text{corner}]$. (b) Forward prediction accuracy of random forests under different maximum tree depths, quantified by mean squared error (MSE). (c) Forward prediction with uncertainty for three testing samples. The uncertainty is quantified by the $\pm 2$ standard deviation denoted as $\pm 2 \hat{s}$. (d) Average uncertainty (quantified by $2 \bar{\hat{s}}$) of dispersion relations across different band orders. }
\label{fig:acoustic_forward}
\end{figure}

Inspired by the micro-architected metamaterial design proposed by Sun et al. \cite{sun2024tailored}, which have been experimentally demonstrated to exhibit a sufficiently rich dynamic property space, we consider a two-dimensional unit-cell design with a similar geometric pattern in this case study, as shown in Fig.~\ref{fig:acoustic_forward}(a). Circular micro-inertia are added to the center and corners of a braced square unit cell with strut radius $r_\text{strut}$. The micro-inertia placed at the center of the brace has radius $r_\text{center}$ while those at the corners of the square unit cell has radius $r_\text{corner}$. The design space $\Omega_\mathbf{x}$ is defined by 4 $\leq r_\text{strut} \leq$ 6.41, $\sqrt{2} r_\text{strut}$ $\leq r_\text{center} \leq$ 20, and $(\sqrt{2}+1) r_\text{strut}$ $\leq r_\text{corner} \leq$ 20 (unit: mm). The lower bounds for $r_\text{center}$ and $r_\text{corner}$ are defined in relation to $r_\text{strut}$ to ensure that the center and corner circles are not completely overlapped by the struts. The unit cell size is set at $l=60$ mm. The Young's modulus $E$, density $\rho$ and Poisson's ratio $\nu$ are set as 70 GPa, 2700 kg/m${^3}$, and 0.33 respectively. To demonstrate the flexibility of requirement specification, we consider design requirements consisting of multiple partial passbands and stopbands (i.e., passbands and stopbands defined over specific wave-vector ranges). 

\subsubsection{Data Acquisition}

The governing equation of the elastic wave propagation can be stated as \cite{wu2023topology, wu2025topology}:
\begin{equation}
\rho \ddot{\mathbf{u}} = \nabla(\lambda+\mu)\nabla\cdot\mathbf{u} + \nabla\cdot \mu \nabla \mathbf{u},
\label{eq:gov}
\end{equation}
where $\rho$ denotes the material density, $\mathbf{u}=\{u_x,u_y\}^\top$ is the displacement vector, $\lambda$ and $\mu$ are the Lamé coefficients. To homogenize an infinite tessellation of an acoustic metamaterial unit cell, the periodic Bloch-Floquet boundary condition is applied:
\begin{equation}
\mathbf{u}(\mathbf{r},\mathbf{k})
= \tilde{\mathbf{u}}(\mathbf{r})\, e^{i(\mathbf{k}^{\mathrm{T}}\mathbf{r}+\omega t)},
\label{eq:bloch}
\end{equation}
where $\mathbf{r}=\{x,y\}$ represents the position vector, $\tilde{\mathbf{u}}$ represents the Bloch displacement vector and $\mathbf{k}=(k_x,k_y)$ is the wave vector. Substituting Eq.~\ref{eq:bloch} into Eq.~\ref{eq:gov} and discretizing based on the finite element method, the following eigenvalue problem can be obtained:
\begin{equation}
\bigl( \mathbf{K}(\mathbf{k}) - \omega^2 \mathbf{M}(\mathbf{k}) \bigr) \, \tilde{\mathbf{u}} = 0,
\label{eq:eig}
\end{equation}
where $\mathbf{K}$ and $\mathbf{M}$ are the Bloch-reduced global stiffness matrix and mass matrix, respectively. Since both $\mathbf{K}$ and $\mathbf{M}$ depend on the wave vector $\mathbf{k}$, the obtained eigenfrequencies $\omega$ are also functions of $\mathbf{k}$.

A total of 500 geometric parameter sets, $\mathbf{x} = (r_\text{strut}, r_\text{center}, r_\text{corner})$, are generated through Latin hypercube sampling within the design space. Bloch-wave analysis is conducted in COMSOL Multiphysics to solve Eq.~\ref{eq:eig} and obtain the dispersion relations. The dispersion relation is computed by sampling wave vectors along the high-symmetry path $\Gamma \rightarrow X \rightarrow M \rightarrow \Gamma$, which traces the boundary of the first irreducible Brillouin zone (IBZ), as illustrated in Fig.~\ref{fig:acoustic_forward}(a). The IBZ boundary is uniformly discretized into 61 points. At each sampled wave vector, the first 15 eigenfrequencies are extracted, yielding a discrete functional response vector $\mathbf{y}$ with dimension $d_{\mathbf{y}} = 915$.

\subsubsection{Forward Prediction}
\label{acoustic fp}

After data acquisition, 400 samples were randomly selected as training data for the random forest to learn the forward mapping, and the remaining 100 samples were held out for testing. Following the reformulated mapping in Eq.~\ref{eq:uncurried}, we use the design variable $\mathbf{x}$ together with the query point $\mathbf{a}=(i,k)$ to predict the corresponding frequency, meaning each component of the response $\mathbf{y}$ is predicted individually. Therefore, the input dimension of the random forest is 5 and the output dimension is reduced to 1. Accordingly, the number of reformulated input--output pairs scales by a factor of $d_{\mathbf{y}}$, resulting in 366,000 pairs for training and 91,500 pairs for testing, respectively. Maximum tree depth is tuned to balance fitting capability and computational cost. As shown in Fig.~\ref{fig:acoustic_forward} (b), when the depth exceeds 20, the mean squared error (MSE, defined as $\frac{1}{d_\mathbf{y}}(\mathbf{y}-\hat{\mathbf{y}})^\top(\mathbf{y}-\hat{\mathbf{y}})$) in both the training and testing sets no longer decreases. Therefore, the maximum tree depth is set to 20, yielding an MSE of 0.0108 for training and 0.4839 for testing.

The forward predictions for three samples from the testing set are shown in Fig.~\ref{fig:acoustic_forward} (c). The blue points represent the mean predictions of the random forest, while the red points denote the ground-truth values obtained from COMSOL simulations. The shaded blue regions indicate the $\pm 2$ standard deviation ranges (denoted as $\pm 2 \hat{s}$) across predictions from individual decision trees, providing a measure of predictive uncertainty as introduced in Sec.~\ref{Forward Prediction with Uncertainty Quantification}. Overall, the ground-truth values are generally well captured within the $\pm 2$ standard deviation bounds. Since higher-order eigenfrequencies generally exhibit greater variability, they are usually more challenging to fit and have a lower prediction accuracy compared to lower-order frequencies \cite{zhang2024deep}. This is captured by the quantified uncertainty, as shown in Fig.~\ref{fig:acoustic_forward} (d). The blue and red solid lines denote the prediction uncertainty across different band orders for the training and testing sets, respectively, quantified as the average of two standard deviation $2 \bar{\hat{s}}$. It is observed that as the band order increases, the standard deviation increases. A similar trend can also be seen in individual predictions in Fig.~\ref{fig:acoustic_forward}(c). This consistency between uncertainty and prediction accuracy indicates that the variance offers a reasonable measure of predictive uncertainty. 

\subsubsection{Inverse Design}

After training the forward model, we perform inverse design using the same model. To evaluate the inverse design performance, 10 partial passband/stopband requirements $\mathcal{T}_k$ were prescribed, as shown in Table~\ref{tab: requirement list}. Each requirement contains 2–3 segments, where each segment specifies a wave-vector range and a frequency range. The wave-vector interval is constructed by randomly selecting the starting point from $\{\Gamma, X, M\}$ and the ending point from $\{X, M, \Gamma\}$, with the constraint that the starting point precedes the ending point along the IBZ boundary. The frequency range is sampled within [20,50] kHz subject to a prescribed bandwidth constraint (1--5 kHz), and non-overlapping frequency intervals are enforced across segments. Once the segments are created, each segment is randomly assigned as either a stopband or a passband. For each requirement, 30 designs are generated using the proposed method by sampling from the corresponding likelihood distribution over the design space. Note that it is possible for the likelihood to be zero everywhere in the design space when the model believes the requirement is unachievable. Such cases are excluded as it is impossible to sample designs from such likelihood distribution.

\begin{table}[h!]
\centering
\footnotesize
\renewcommand{\arraystretch}{1.2}

\caption{Design requirements used for inverse design. Each band is specified by a wave-vector range and a frequency range (unit: kHz). }

\begin{tabularx}{0.98\linewidth}{
    c|
    >{\centering\arraybackslash}X|
    >{\centering\arraybackslash}X
}
\hline
\textbf{Requirement} & \textbf{Stopbands} & \textbf{Passbands} \\ \hline

$\mathcal{T}_1$ &
$\left[\Gamma\!-\!X\right], [45.48,\,47.52]$ &
$\left[X\!-\!M\right], [37.49,\,39.12]$ \\ \hline

$\mathcal{T}_2$ &
\makecell{$\left[M\!-\!\Gamma\right], [26.20,\,27.94]$\\
$\left[X\!-\!M\!-\!\Gamma\right], [47.85,\,49.91]$} &
$\left[M\!-\!\Gamma\right], [22.83,\,25.49]$ \\ \hline

$\mathcal{T}_3$ &
\makecell{$\left[X\!-\!M\!-\!\Gamma\right], [26.57,\,28.22]$\\
$\left[\Gamma\!-\!X\right], [21.91,\,24.34]$} &
$\left[X\!-\!M\right], [45.27,\,47.46]$ \\ \hline

$\mathcal{T}_4$ &
$\left[\Gamma\!-\!X\right], [47.04,\,48.39]$ &
$\left[\Gamma\!-\!X\right], [43.09,\,45.96]$ \\ \hline

$\mathcal{T}_5$ &
$\left[M\!-\!\Gamma\right], [45.20,\,46.21]$ &
$\left[\Gamma\!-\!X\!-\!M\!-\!\Gamma\right], [28.42,\,30.15]$ \\ \hline

$\mathcal{T}_6$ &
$\left[M\!-\!\Gamma\right], [44.96,\,47.17]$ &
\makecell{$\left[\Gamma\!-\!X\right], [26.87,\,28.72]$\\
$\left[X\!-\!M\right], [32.96,\,34.03]$} \\ \hline

$\mathcal{T}_7$ &
\makecell{$\left[M\!-\!\Gamma\right], [26.95,\,28.12]$\\
$\left[\Gamma\!-\!X\!-\!M\right], [32.14,\,33.41]$} &
None \\ \hline

$\mathcal{T}_8$ &
\makecell{$\left[M\!-\!\Gamma\right], [27.66,\,28.82]$\\
$\left[\Gamma\!-\!X\!-\!M\right], [35.54,\,38.05]$} &
None \\ \hline

$\mathcal{T}_9$ &
\makecell{$\left[\Gamma\!-\!X\!-\!M\right], [37.71,\,40.34]$\\
$\left[X\!-\!M\right], [24.94,\,26.90]$} &
None \\ \hline

$\mathcal{T}_{10}$ &
\makecell{$\left[\Gamma\!-\!X\right], [30.37,\,33.22]$\\
$\left[\Gamma\!-\!X\right], [46.33,\,48.07]$} &
None \\ \hline

\end{tabularx}
\label{tab: requirement list}
\end{table}

\begin{figure}[!h]
\centering
\includegraphics[width=1\textwidth]{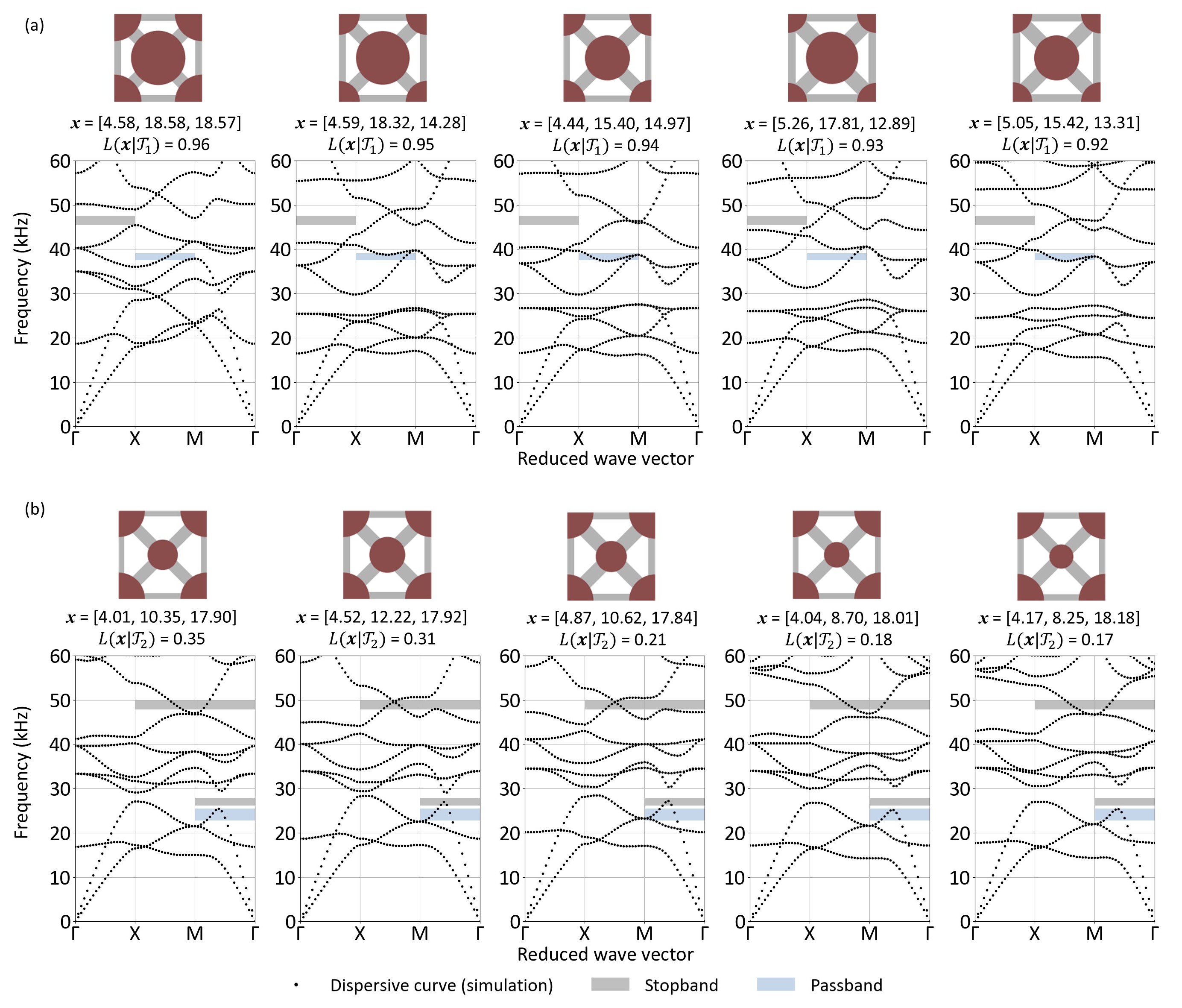}
\caption{Inverse design results for acoustic metamaterials for two different requirements $\mathcal{T}_1$ (a) and $\mathcal{T}_2$ (b). The blue shaded area indicates the partial passband and the gray shaded area indicates the partial stopband. For each requirement, the top-5 highest likelihood unit cell geometries and the corresponding dispersion relations are displayed. For $\mathcal{T}_1$, all the five designs meet the requirement. For $\mathcal{T}_2$, none meets the requirement. The detailed specifications of all the ten requirements can be found in Table~\ref{tab: requirement list}.}
\label{fig:acoustic_inverse_case}
\end{figure}

Fig.~\ref{fig:acoustic_inverse_case} (a) and (b) show the designs generated through RAG for requirements $\mathcal{T}_1$ and $\mathcal{T}_2$, respectively. The blue shaded area denotes the target passband and the gray shaded area denotes the target stopband. For each requirement, we present the five highest-likelihood designs. For $\mathcal{T}_1$, which imposes relatively few constraints, feasible designs are easy to identify, thus all five designs generated by RAG attain high likelihood (all higher than 0.90) and all satisfy the requirement as confirmed by simulation. Since many dispersion relations can satisfy these constraints, the corresponding design solutions are distributed across multiple regions of the design space. This characteristic is also well captured by RAG, as the generated unit-cell geometries and their corresponding dispersion relations are clearly distinct. This demonstrates that RAG effectively handles the one-to-many nature of the inverse problem by producing diverse solutions for a single requirement. In comparison, for $\mathcal{T}_2$ that introduces more constraints, the likelihood of the generated designs is relatively low (all lower than 0.4) and none of them satisfy the requirement. Overall, likelihood correlates strongly with feasibility: higher-likelihood designs are more likely to meet the requirement. Leveraging this information allows us to assess the complexity of a given inverse-design requirement and mitigate risk by excluding low-likelihood designs.

\begin{figure}[!h]
\centering
\includegraphics[width=1\textwidth]{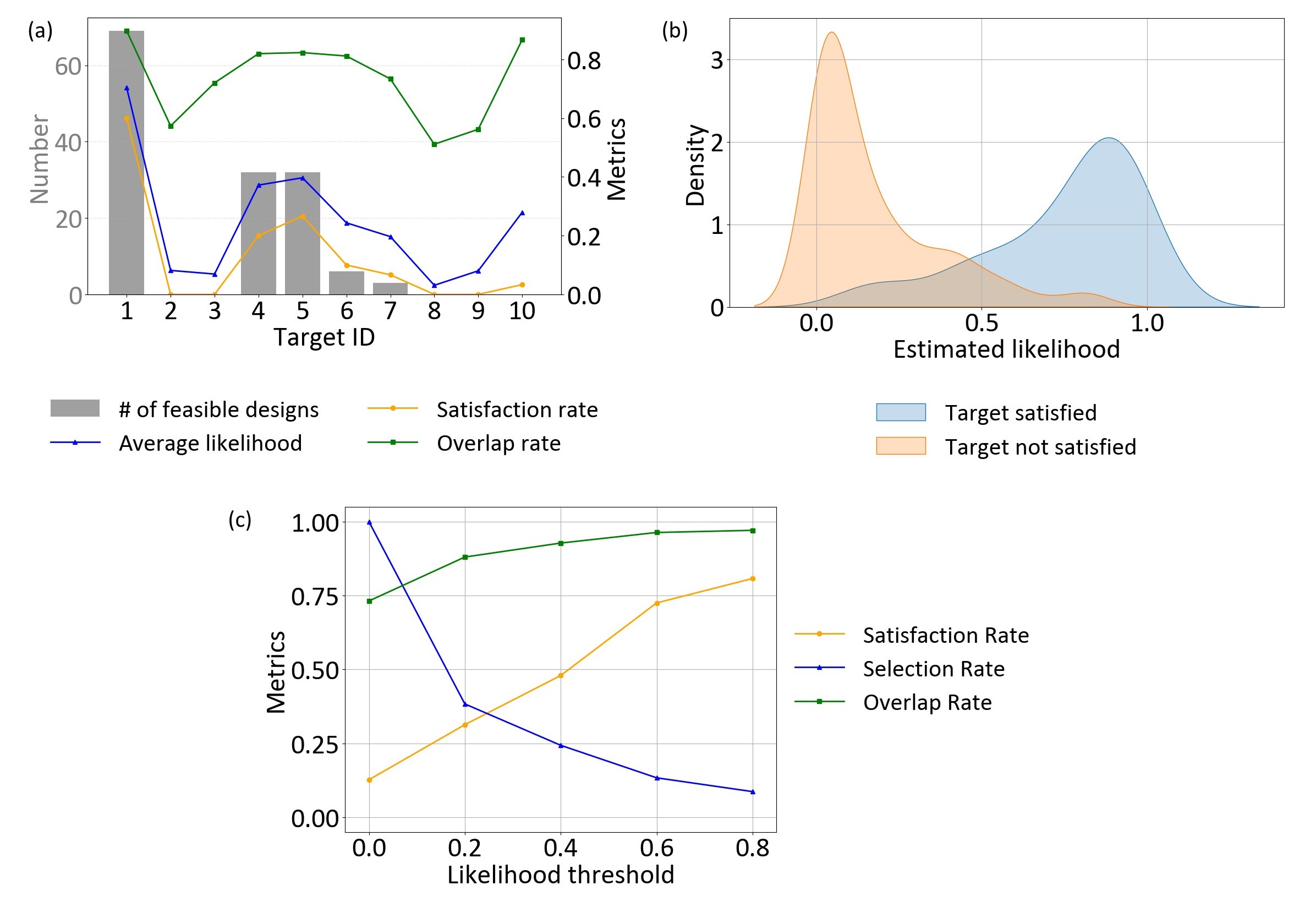}
\caption{Statistical summary of all 300 designs generated from the 10 prescribed partial passband/stopband requirements in Table~\ref{tab: requirement list}. (a) The overview of the 10 requirements. The gray bar represents the number of training samples satisfying their corresponding requirements. The blue, yellow, and green curves represent the average likelihood, satisfaction rates, and average overlap rates, respectively. (b) KDE of the estimated likelihood for all 300 generated designs. (c) Satisfaction rates, overlap rates, and selection rates for generated designs under varying sampling thresholds.}
\label{fig:acoustic_inverse_stats}
\end{figure}

Fig.~\ref{fig:acoustic_inverse_stats} (a) shows an overview of all 10 prescribed requirements. The gray bars denote the number of training samples satisfying each requirement. In general, the more feasible samples present in the dataset, the easier it is to achieve the corresponding requirement within the design space. Accordingly, requirements such as $\mathcal{T}_1$, $\mathcal{T}_4$, and $\mathcal{T}_5$ are relatively easy to achieve, while $\mathcal{T}_2$, $\mathcal{T}_3$, $\mathcal{T}_8$, and $\mathcal{T}_9$ are more difficult because no feasible samples exist in the training dataset. This difficulty is effectively captured by the average likelihood shown in the blue curve: requirements that are easier to achieve exhibit higher average likelihoods, while those lacking feasible samples show lower values. This trend is consistent with Fig.~\ref{fig:acoustic_inverse_case}, confirming that likelihood serves as an effective indicator of requirement difficulty.

To show that RAG can produce designs that satisfy the requirement, we compute the satisfaction rate---defined as the fraction of generated designs meeting the requirement (yellow curve)---and the average overlap rate, which measures the average frequency range overlap between a generated response and the requirement. The satisfaction rate is computed based on a binary measure (a design either satisfies the full requirement or not), while the overlap rate offers a softer measure that accounts for partial satisfaction. By construction, the average overlap rate is lower-bounded by the satisfaction rate, revealing when the model generates designs that, although infeasible, remain close to the requirement. For relatively easy requirements ($\mathcal{T}_1$, $\mathcal{T}_4$, $\mathcal{T}_5$), both satisfaction and overlap rates are high, whereas for difficult requirements ($\mathcal{T}_2$, $\mathcal{T}_3$, $\mathcal{T}_8$, $\mathcal{T}_9$) both measures are low. Importantly, satisfaction and overlap rates can only be obtained through explicit feasibility validation. In contrast, our framework leverages likelihood information to flag designs that are less likely to meet the requirement prior to validation.

Fig.~\ref{fig:acoustic_inverse_stats} (b) shows the kernel density estimation (KDE) of the likelihood for all 300 generated designs, conditioned on their corresponding requirements. It can be observed that infeasible designs generally exhibit low likelihood, whereas feasible designs are associated with high likelihood values. Therefore, by applying a likelihood threshold to filter out low-likelihood designs, the overall satisfaction rate of the generated designs can be increased. The relationships among the selection rate, satisfaction rate, overlap rate, and the likelihood threshold are shown in Fig.~\ref{fig:acoustic_inverse_stats}(c). As the likelihood threshold increases and low-likelihood designs are progressively filtered out, the selection rate decreases, while the remaining designs exhibit increasing satisfaction and overlap rates. This also demonstrates that the likelihood threshold can be flexibly adjusted according to the acceptable confidence level of different tasks.

\subsection{Design Mechanical Metamaterials with Desired Snap-Through Response}
\label{Design Mechanical Metamaterials with Target Snap-Through Response}

\subsubsection{Problem statement}

Mechanical metamaterials have been widely studied due to the unprecedented mechanical properties that arise from their unit-cell structures \cite{jiao2023mechanical}. These properties are often characterized by their stress--strain relations  \cite{chai2024tailoring, deng2022inverse, he2025customizable, ha2023rapid}, which can be represented as a function $\sigma(\epsilon)$, with the query point $\mathbf{a} := \epsilon$ representing the strain. To demonstrate both the capability of RAG for stress--strain relation inverse design and its data-efficiency advantage, we apply it to an existing mechanical metamaterial system proposed by Chai et al. \cite{chai2024tailoring}, using only 40\% of the original training dataset. 

\begin{figure}[!h]
\centering
\includegraphics[width=1\textwidth]{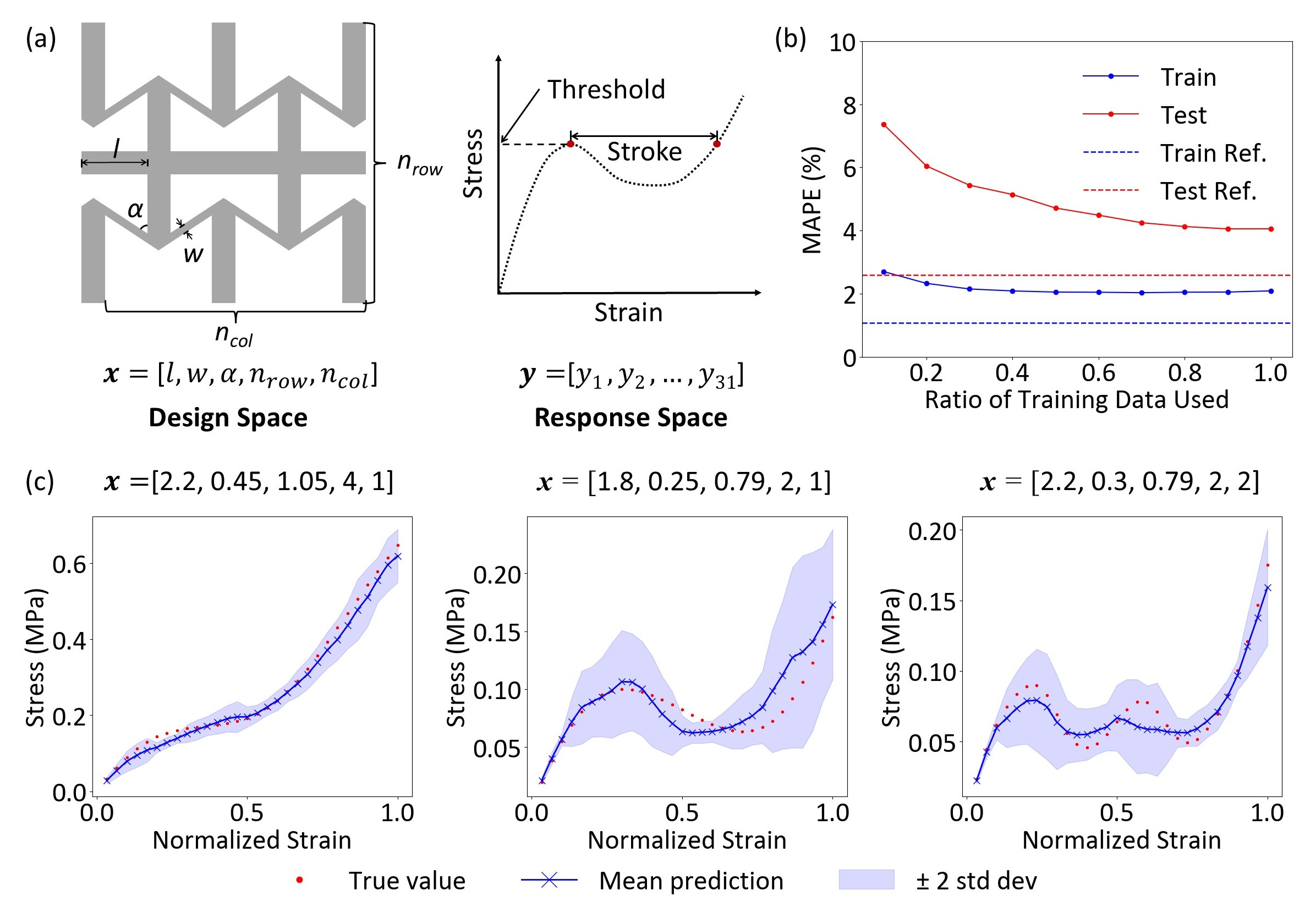}
\caption{Design problem statement and forward modeling of mechanical metamaterials. (a) Design space and response space in mechanical metamaterials. (b) Relationship between prediction accuracy, quantified by mean absolute percentage error (MAPE), and the ratio of data used for model training. The solid blue and red lines represent the training and testing accuracy of the random forest model, respectively. The dashed blue and red lines represent the training and testing accuracy of the NN reported in Ref.~\cite{chai2024tailoring}. (c) Forward prediction and uncertainty quantification for test samples without snap-through, with one snap-through wave, and with two snap-through waves, respectively. }
\label{fig:mechanical_forward}
\end{figure}

In this task, the design space is defined by five variables: the horizontal length of the tilt beam $l$, the angle between the cantilever beam and tilt beam $\alpha$, the width of the tilt beam $w$, and the number of columns $n_{col}$ and rows $n_{row}$, as shown in Fig.~\ref{fig:mechanical_forward} (a). $n_{col}$ and $n_{row}$ can only take integer values of 1, 2, 3, or 4. The stress--strain relation is uniformly discretized into 31 points; since the stress at the first point is always zero, we predict only the remaining 30 points, giving the discrete functional response vector $\mathbf{y}$ a dimensionality of 30. In this study, we focus on the reconfigurable soft actuator inverse design task introduced in \cite{chai2024tailoring}, where different actuation modes can be achieved under a simple pneumatic input. The key to achieve this functionality is to find unit cell design that can snap through under specific force and provide desired stroke. Therefore, the design requirement imposed on the stress--strain relation is specified in terms of a target snap-through threshold force and stroke, as illustrated in Fig.~\ref{fig:mechanical_forward}(a).

\subsubsection{Forward modeling}

Similar to the forward prediction in the acoustic case (Sec.~\ref{acoustic fp}), we jointly take the design variable $\mathbf{x}$ and the query point $\mathbf{a}$—--which corresponds to the strain $\epsilon$ in this task—--as inputs to predict the corresponding stress value. A random forest regressor with the maximum tree depth of 15 is employed for this task. The public ground dataset consists of 2065 training samples and 231 testing samples. To evaluate the data efficiency of RAG, we downsample the original training set with varying downsampling portions, while keeping the testing set unchanged. Each subset is then used to train a separate random forest model and the corresponding accuracy in shown in Fig.~\ref{fig:mechanical_forward} (b). The results indicate that random forests remain effective with limited data: when only 10\% of the training data are used, the model still achieves mean relative errors of 2.70\% and 7.37\% on the training and testing sets, respectively. In comparison, the original study using NN achieved lower errors of 1.08\% (training) and 2.59\% (testing), but required the full dataset. This result shows that, the proposed RAG framework can attain prediction accuracy close to that of NN with a significantly smaller amount of training data, thereby highlighting its data-efficiency advantage. For subsequent tasks, we select the model trained on 40\% of the training data (826 samples), which achieves 2.09\% error on the training set and 5.14\% on the testing set.

Fig.~\ref{fig:mechanical_forward} (c) shows the forward predictions with uncertainty quantification for test samples without snap-through, with one snap-through wave, and with two snap-through waves, respectively. It can be observed that the curves with greater fluctuation exhibit wider uncertainty ranges. The ground truth values remain within the $±2$ standard deviation bands, demonstrating both reasonable predictive accuracy and meaningful uncertainty quantification.

\subsubsection{Inverse inference}

The design requirement is specified as a target threshold force of 0.2 MPa and a target stroke of 1.3 mm. We relax the requirement by introducing a tolerance on these values---that is, the generated designs are allowed to have threshold force and stroke values within a specified percentage range of the targets. Requirements with smaller tolerances are more difficult to satisfy. Here, we consider two different tolerance levels, ±25\% and ±15\%, to investigate how the difficulty of the requirement influences the estimated likelihood.

\begin{figure}[!h]
\centering
\includegraphics[width=1\textwidth]{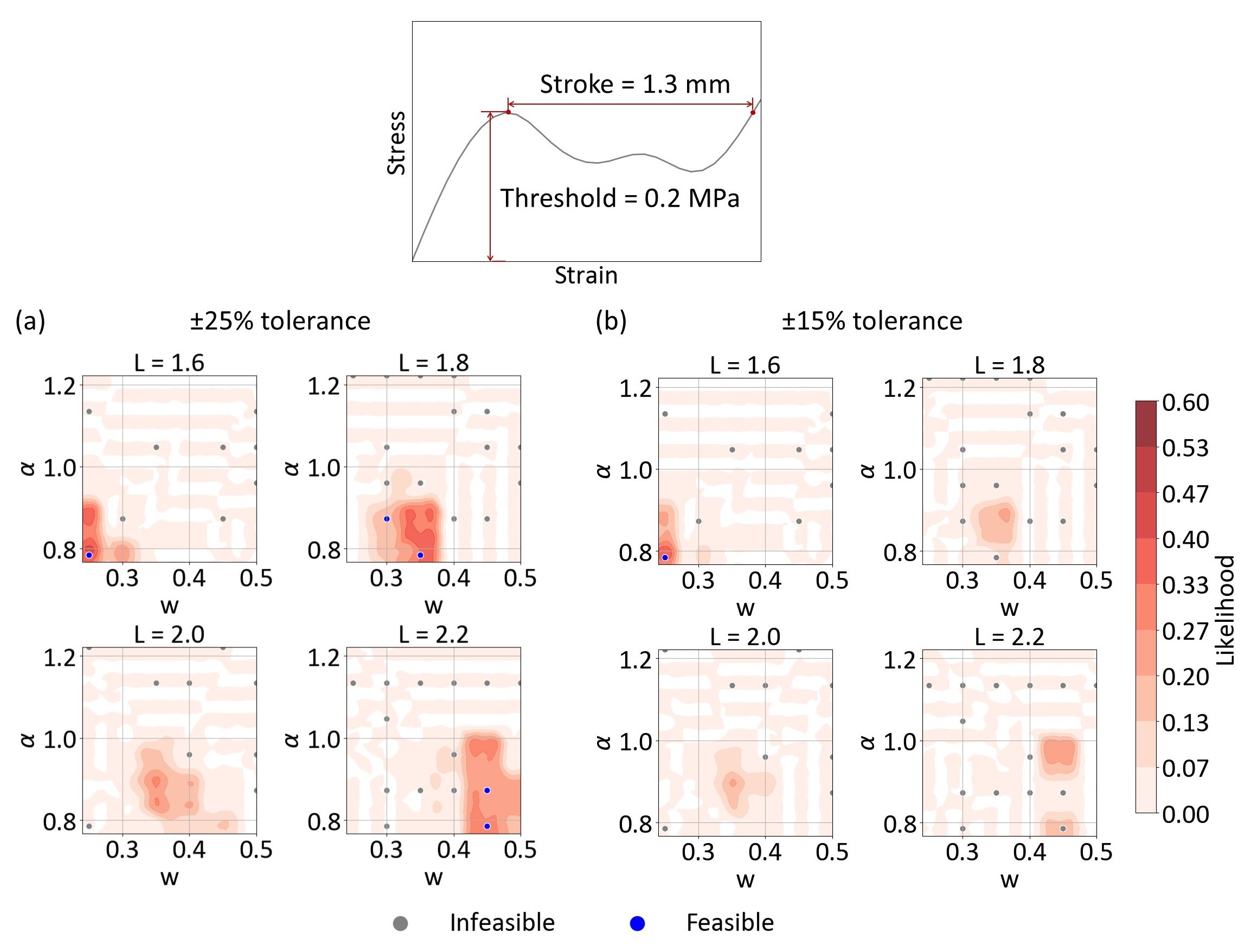}
\caption{Likelihood distribution in the design space for desired stroke and threshold with different tolerances. The gray/blue points denote the infeasible/feasible data points in the training dataset (a) tolerance = 25\% (b) tolerance = 15\%.}
\label{fig:mechanical_inverse}
\end{figure}

Fig.~\ref{fig:mechanical_inverse} (a) and (b) visualizes the estimated likelihood distribution over the design space for tolerance levels of ±25\% and ±15\%. The discrete design variables $n_{col}$ and $n_{row}$ divide the 5-dimensional design space into sixteen 3-dimensional subspaces. Here we only visualize the subspace corresponding to $n_{col} = 4$ and $n_{row} = 2$ as an example. To examine the accuracy of the estimated likelihoods, we also plot the training samples observed by the random forests (i.e. the reduced training dataset). Gray points represent samples that do not meet the requirement, while blue points denote those that satisfy it. It can be observed that regions with high estimated likelihoods are located near feasible samples, whereas low likelihoods appear in regions surrounded by infeasible ones. Since the requirements on threshold and stroke do not impose constraints on the specific shape of the stress-–strain relation, different types of stress–-strain relations may satisfy these requirements. Consequently, multiple high-likelihood regions can be observed across the design space. Moreover, as the tolerance becomes tighter, the overall likelihood decreases while the landscape of the likelihood distribution remains similar because of the unchanged target. These observations suggest that the likelihood distribution estimated by RAG can effectively reflect the distribution of feasible design solutions, and that the magnitude of the likelihood also serves as a reliable indicator of the difficulty of the design requirement.

\section{Discussion}
\label{Discussion}

In this section, we discuss the advantages of RAG in terms of flexibility in requirement specification, data efficiency, and uncertainty awareness. We further analyze its computational efficiency to illustrate how the problem structure influences the overall computational performance of RAG.

\subsection{Flexibility in Requirement Specification}

Existing generative models often learn the inverse mapping from the functional response $\mathbf{y}$ to the design $\mathbf{x}$. Therefore, representative functional responses that satisfy the design requirements (e.g., stress--strain relations with desired threshold and stroke, or dispersion relations with on-demand bandgap property) must be manually specified as inputs to generate the corresponding design solutions. In contrast, RAG directly learns the inverse mapping from the design requirement $\mathcal{T}$ to the design $\mathbf{x}$ by estimating the likelihood conditioned on the requirement, i.e., $\mathcal{L}(\mathbf{x} \mid \mathcal{T})$. This eliminates the need to manually specify representative functional responses for design generation and enables greater flexibility in requirement specification.

In the two design tasks considered in this work---partial passband/stopband control (Sec.~\ref{Design Acoustic Metamaterials with On-Demand Partial Passbands/Stopbands}) and snap-through threshold/stroke control (Sec.~\ref{Design Mechanical Metamaterials with Target Snap-Through Response})---the requirements do not prescribe the exact shape of the functional response and thus allow functional responses with diverse shapes. In such cases, it is challenging to formulate the requirements as differentiable objectives amenable to topology optimization, and identifying representative dispersion or stress–strain relations for generative models is also nontrivial. Moreover, using a specific representative functional response to approximate the requirement would inevitably introduce bias and artificially restrict the feasible design space. In contrast, RAG avoids this source of constraint and enables a more comprehensive exploration of feasible designs across the design space.

\subsection{Data Efficiency}

Our method enables forward prediction and inverse design of full dispersion relations and stress--strain relations using significantly fewer data than those required by existing NN-based approaches for similar tasks. The data efficiency of RAG mainly arises from two factors. First, random forests are inherently more robust than NN in small-data regimes. Second, instead of treating the entire high-dimensional functional response as the model output, we introduce the query points of the functional response as model inputs and reduce the model output dimension to one. This reformulation substantially lowers the learning complexity, thereby improving data efficiency.

\subsection{Uncertainty Awareness}

By leveraging the ensemble structure of random forests, our framework can quantify uncertainty both in the response space during forward prediction and in the design space for arbitrary requirements during inverse design. This capability is particularly critical for inverse design problems, which can be either one-to-many---where the requirements are relatively easy and multiple design solutions satisfy the same requirement---or one-to-none---where the requirements are too stringent to allow any feasible designs within the predefined design space. Most existing generative inverse design frameworks produce design solutions without providing any measure of the conditional inference. As a result, even when the requirement is challenging or even impossible to satisfy, the approaches provide no ways to monitor this. In contrast, our method provides model uncertainty information that helps indicate how difficult a given requirement is to satisfy within a predefined design space. The trustworthiness of the generated design solutions can then be improved by selecting those with higher confidence (likelihood) values.

\subsection{Computational Efficiency and Scalability}

The applicability of random forests is constrained by their computational efficiency. The time complexity of training a random forest is $O(NMk\log(M))$, where $N$ is the number of trees, $M$ is the number of input--output pairs, and $k$ is the number of features considered at each split. During data acquisition, we often compute the functional response $\mathbf{y}$ for a given design $\mathbf{x}$. Therefore, the number of training design samples, denoted as $m$, corresponds to the number of distinct $\mathbf{x}-\mathbf{y}$ pairs. Based on the reformulation in Eq.~\ref{eq:uncurried}, the query point $\mathbf{a}$ is introduced as additional inputs to predict the corresponding functional response value. As a result, each original $\mathbf{x}$ - $\mathbf{y}$ pair is decomposed into $d_\mathbf{y}$ input--output pairs. Under this setting, we have $M=m\sqrt{d_\mathbf{y}}$ and $k=\sqrt{d_\mathbf{x}+d_\mathbf{a}}$. The time complexity for training can be then expressed as \cite{louppe2014understanding}
\begin{equation}
O(Nm\sqrt{d_\mathbf{y}(d_\mathbf{x}+d_\mathbf{a})}\log(m\sqrt{d_\mathbf{y}}))
\label{eq:complexity}
\end{equation}
This scaling relation illustrates how the computational efficiency of the random forest depends on the dimensionality of the functional response and the sample size. Based on Eq.~\ref{eq:complexity}, one can evaluate the trade-off between the resolution of the functional response determined by $d_\mathbf{y}$ and the training efficiency. For reference, the actual training time for each scenario in this study is summarized in Table~\ref{tab:training_time}, obtained on an Intel Core i7-12700 CPU (2.10 GHz) with 32 GB of memory. It can be observed that both cases are trained within one minute, which is substantially faster than the NN-based approach. The acoustic metamaterial case requires a longer training time mainly because the dispersion relation has a much higher dimensionality than the stress-–strain relation.
\begin{table}[htbp]
\centering
\caption{Training time of random forests in each task.}
\label{tab:training_time}
\begin{tabular}{lcccc}
\hline
\textbf{Task} & $d_\mathbf{y}$ & $m$ & $d_\mathbf{x}$ & \textbf{Training time (s)} \\ 
\hline
Acoustic metamaterials design & 915  & 400 & 3 & 58.88 \\
Mechanical metamaterials design & 30  & 826 & 5 & 3.01 \\
\hline
\end{tabular}
\end{table}
In the design generation stage, the efficiency of MCMC sampling is primarily influenced by the dimensionality of the design space $d_\mathbf{x}$. A high-dimensional design space makes it more difficult for the MCMC process to explore and locate regions with high likelihood. When the requirement is challenging to achieve, the corresponding high-likelihood region may be extremely small or even nonexistent, which further increases the sampling time. To ensure computational efficiency, the proposed method is therefore more suitable for parametric design spaces with relatively low dimensionality.

\section{Conclusion}
\label{Conclusion}

In this work, we presented RAG, a random-forest-based generative inverse-design framework capable of handling quantitative, high-dimensional functional responses using small datasets. RAG requires training only a single forward random-forest model, which avoids extensive hyperparameter tuning and enables fast, stable training. Instead of directly learning the mapping from design to functional response, we take design variables $\mathbf{x}$ together with the query point $\mathbf{a}$ of the functional response as the input and learn their joint mapping. This reformulation reduces the learning difficulty by lowering the output dimensionality of the model and enables discretization-invariant prediction. During inverse design, RAG directly estimates the likelihood conditioned on the requirement, providing substantial flexibility in how design requirements are specified. Leveraging the random forests ensemble, the framework quantifies uncertainty in both forward prediction---through the variance across tree outputs---and inverse design---through the estimated likelihood. These uncertainty measures reflect the model’s confidence in its inferences. Iteration-free, single-shot design generation can be achieved by sampling the design space from the estimated likelihood distribution.

We validated RAG on two metamaterial systems: acoustic metamaterials with prescribed partial passbands/stopbands, and mechanical metamaterials with desired snap-through behaviors, where the design requirements are challenging to address using gradient-based topology optimization. In both cases, RAG operates effectively with far fewer data than those required by existing deep-learning-based methods, while accurately predicting high-dimensional responses and producing diverse, feasible design candidates. The likelihood estimates can help secure trustworthiness of the generated design by filtering low-confidence solutions and indicate requirement difficulty.

In addition, we highlight two potential directions for future work that can further exploit the uncertainty-aware and easy-to-construct advantages of RAG. First, the uncertainty information provided by RAG in forward prediction can be leveraged for adaptive sampling to improve data efficiency. By selectively adding new design samples in high-uncertainty regions, the total number of required samples can be significantly reduced while maintaining prediction accuracy. It is worth noting that RAG learns a mapping from the joint space of design variables and query points of the functional response. Therefore, adaptive sampling can be simultaneously applied in the design space $\Omega_{\mathbf{x}}$ and the query-point space $\Omega_{\mathbf{a}}$. This implies that the discretization of the functional response itself can also be optimized. Second, owing to its ease of construction, RAG can also be used to quickly assess whether a user-defined design space is sufficiently rich to satisfy a given design requirement. In practice, it is often unclear whether the predefined design space is adequate for achieving the requirement. RAG provides an efficient tool for such an examination. If, for a given requirement, the conditional likelihood distribution estimated by RAG is generally low across the entire design space, this indicates that the predefined design space is insufficient and should be expanded to meet the requirement.


Overall, RAG offers a lightweight, uncertainty-aware, and data-efficient pathway to metamaterial inverse design with nonlinear and high-dimensional functional responses. Its flexibility in requirement specification, robustness in small-data regimes, and compatibility with adaptive learning strategies make it a promising building block for uncertainty-informed metamaterials design pipelines. Beyond the metamaterials design demonstrated in this work, RAG can be applied to more general inverse-design problems involving high-dimensional property spaces, expensive physics-based simulations, and complex functional requirements, possibly upon trivial modifications.




\section*{Acknowledgments}

Wei Chen, Doksoo Lee, and Bolin Chen acknowledge support from the NSF Boosting Research Ideas for Transformative and Equitable Advances in Engineering (BRITE) Fellow Program (CMMI-2227641). Wei Chen and Doksoo Lee are also grateful for partial support from NASA's Minority University Research and Education Project Institutional Research Opportunity (MIRO) through the Center for In-Space Manufacturing (CISM-R2): Recycling and Regolith Processing (Award 80NSSC24M0176). Wei "Wayne" Chen acknowledges support from the NSF EDSE program (CMMI 2434393). The authors also acknowledge Dr. Mohammad Charara, Rachel Sun, and Prof. Carlos M. Portela for insightful discussions and constructive comments on the manuscript.

\section*{Data Availability Statement}

The source code and data for the acoustic metamaterial case in this work are available at: \url{https://github.com/Sautoy/Random_Forest_Based_Generative_Design_Framework}.

\bibliographystyle{unsrt}  
\bibliography{references}  

\end{document}